\documentclass[preprint,12pt]{elsarticle}

\usepackage{amssymb}
\usepackage{amsmath}

\usepackage[labelformat=simple]{caption}
\usepackage{amsmath,amssymb,amsfonts}
\usepackage{tabularx}
\usepackage[ruled,vlined]{algorithm2e}
\usepackage{graphicx}
\usepackage{float}
\usepackage{hyperref}
\usepackage{booktabs}
\usepackage{textcomp}
\usepackage{xcolor}
\usepackage{adjustbox}
\usepackage[justification=centering]{caption}

\hypersetup{
    colorlinks=true,
    linkcolor=blue
    }
\journal{Data \& Knowledge Engineering}

\begin{document}

\begin{frontmatter}



\author{Jiabin Xue\corref{cor1}\fnref{first-author}}
\ead{jia22345678@gmail.com}
\fntext[first-author]{Jiabin Xue is the first author}
\cortext[cor1]{Corresponding author}

\title{Semi-Supervised Online Learning on the Edge by Transforming Knowledge from Teacher Models}


\affiliation{organization={Department of Computer Science and Information Science, University of Strathclyde},
            city={Glasgow},
            postcode={G1 1XQ}, 
            country={United Kingdom}}

\begin{abstract}
Edge machine learning (Edge ML) enables training ML models using the vast data distributed across network edges. However, many existing approaches assume static models trained centrally and then deployed, making them ineffective against unseen data. To address this, Online Edge ML allows models to be trained directly on edge devices and updated continuously with new data. This paper explores a key challenge of Online Edge ML: “How to determine labels for truly future, unseen data points”. We propose Knowledge Transformation (KT), a hybrid method combining Knowledge Distillation, Active Learning, and causal reasoning. In short, KT acts as the oracle in active learning by transforming knowledge from a teacher model to generate pseudo-labels for training a student model. To verify the validity of the method, we conducted simulation experiments with two setups: (1) using a less stable teacher model and (2) a relatively more stable teacher model. Results indicate that when a stable teacher model is given, the student model can eventually reach its expected maximum performance. KT is potentially beneficial for scenarios that meet the following circumstances: (1) when the teacher’s task is generic, which means existing pre-trained models might be adequate for its task, so there will be no need to train the teacher model from scratch; and/or (2) when the label for the student’s task is difficult or expensive to acquire.
\end{abstract}


\end{frontmatter}


\section{Introduction}\label{intro}
The integration of ML and IoT raises the insight that “Machine learning can bring intelligence to IoT applications” \cite{ref1}. Traditionally, such integration was achieved by remotely centralised ML, which in practice the ML models are deployed in the cloud or on servers, where IoT devices on the edge are responsible for capturing data and then sending it via the internet, so the model is trained remotely. Once the model is completely trained, it can be used to make inferences for new data, where inferences are carried out on the cloud and then predictions are sent to edge IoT devices.

The traditional manner succeeded in exploiting the vast computational power of centre points (e.g., cloud and servers) to explore data that was extensively scattered at the edge. However, this approach is vulnerable in several aspects, such as latency, security, and privacy because it requires frequent data transmissions between the edge devices and the centre points. Therefore, a new paradigm called Edge ML emerged to utilise the computational power of edge IoT devices to take over partial or all computational tasks (i.e., training and/or inference), which allows data to stay locally. A realistic challenge for Edge ML is the specification of edge IoT devices because most IoT devices deployed at the edge are embedded with Microcontroller Units (MCUs), a circuit board with limited resources in processing speed and memory storage compared to general computers, e.g., a few MHz to several hundred MHz of clock speed and less than 1 MB of memory. 

Existing approaches for Edge ML: Federated learning (FL) introduced the way of training ML models by gathering numerous edge IoT devices together and distributing the global model onto each of them for training. This method effectively utilised the computational power of edge devices; however, it still struggles with the privacy and latency concerns that the traditional method encountered. 

Another approach is similar to the traditional method. The conventional way of this approach is that it pretrains ML models on the cloud/server first, then it deploys the pretrained model on edge IoT devices directly. This allows IoT devices to be capable of making inferences on-device, and there is no need to send data over long distances, which effectively reduces the risk of privacy and latency. This method is further optimised by compression techniques, such as weight pruning, weight sharing, quantisation, et cetera. However, this approach does not guarantee that the pretrained model will perform well on real unseen data. If we expect the pretrained model to be powerful enough, then we need to add more training data and to increase model complexity. For the former, there might be no more training data because labelling data is expensive, whereas for the latter, a larger model (even after optimisations) might not be affordable for edge devices to carry out the inference task. Also, the computational power of edge devices is not fully utilised since it is inference-only.

Online training on edge devices has shown its potential for Edge ML. Since the model is trained and used locally, there is neither latency nor security concern, and it is friendly to MCU-based devices because we can train one data point at a time, which greatly reduces the requirement for processing speed and memory capacity. 

\begin{figure}
\centering
\includegraphics[width=0.95\linewidth]{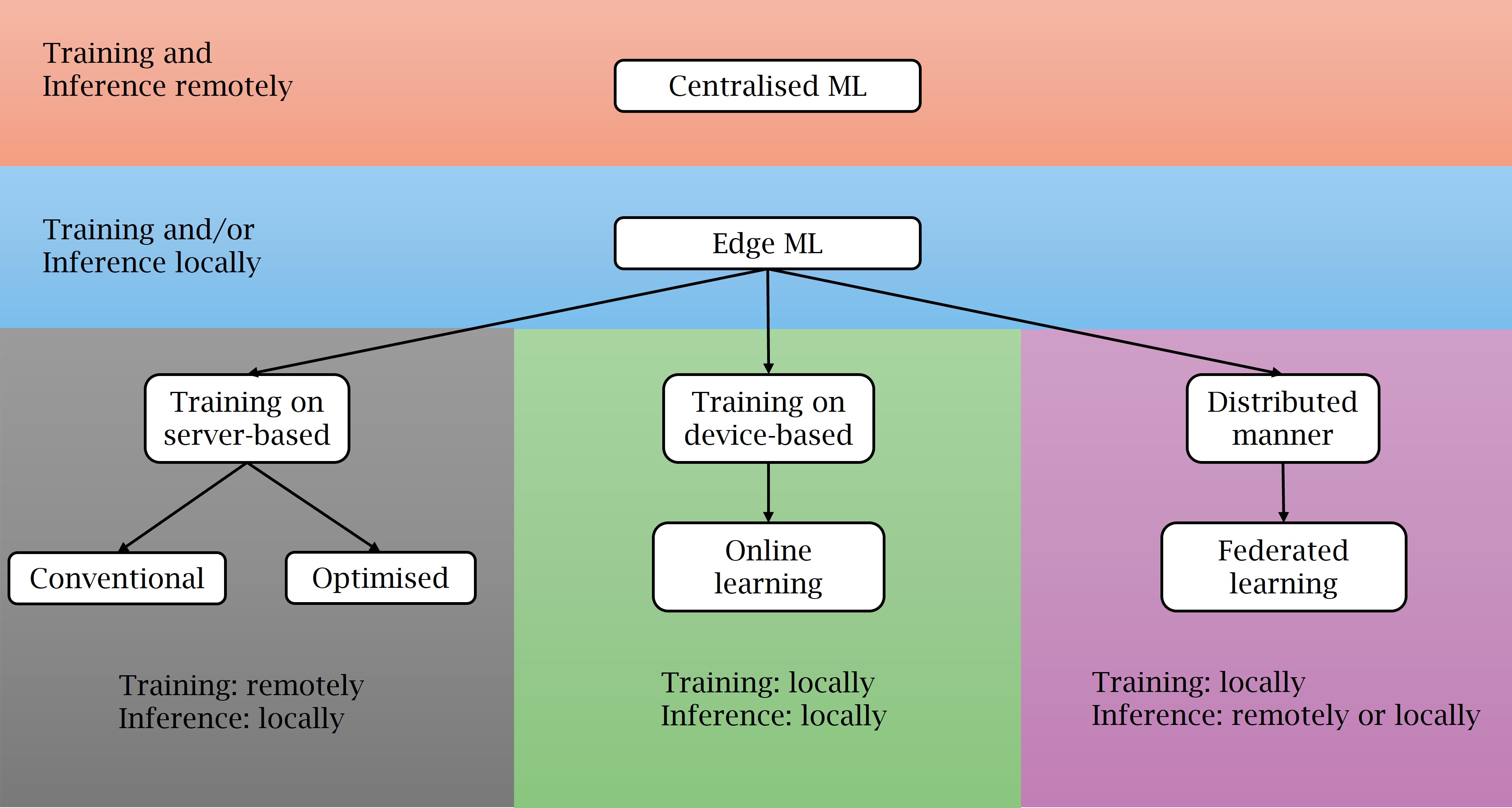}
\caption{\label{fig:Edge ML approaches}Traditional approaches vs Current Edge ML approaches}
\end{figure}

To train ML models in an online-learning manner on real edge scenarios, there is one question we must answer: “What is the label of future data?”. Because without labels provided, supervised ML models cannot be trained via backpropagation. Existing solutions to online learning on edge devices assume that there are labels provided for future data but do not mention how. In fact, this question is crucial if an online-learning model is practically deployed on the edge and aimed at continuously training on new data, but the data target is unknown.

\begin{figure}
\centering
\includegraphics[width=0.95\linewidth]{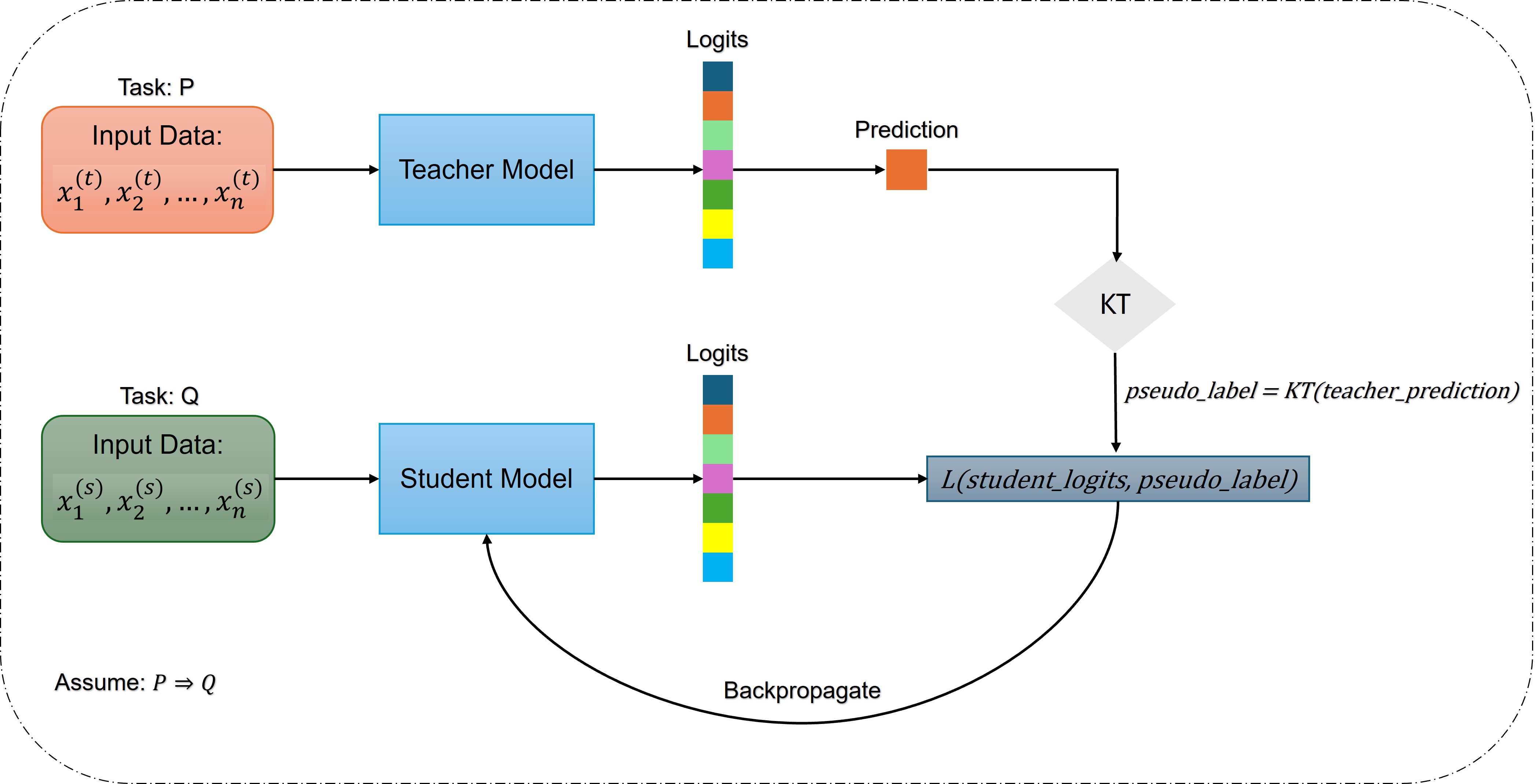}
\caption{Overview of the proposed method. KT relies on the assumption that there exist certain causal relationships between the teacher's task and the student's task (e.g., $P \Rightarrow Q$), and it does not force statistical relations between the data distribution of both models inputs.}
\label{fig:KT_Overview}
\end{figure}

To acquire labels for truly unseen data, we propose a hybrid approach based on Knowledge Distillation (KD) \cite{ref2} and Active Learning (AL) \cite{ref34}. The entire system involves using two models, the student and the teacher, where the student is the model that performs the main task in scenarios (e.g., classification, object detection, regression), then the teacher model is responsible for providing its knowledge, and both models are probable to be deployed at the edge directly. Different from traditional KD, the proposed approach does not require both student and teacher to use the same dataset, and hence the student does not have to mimic the teacher's behaviour as its objective. Instead of distillation, the proposed method is achieved by transforming knowledge from the teacher model based on certain causal reasoning between the tasks of both models, and the transformed knowledge can be treated as the (pseudo-)label used for training the student model. Therefore, this method can be seen as a variant of KD, and we describe it as Knowledge Transformation (KT). Note that this work is primarily theoretical, and the experiments are the simulation of the process of running the proposed method in given scenarios and verifying its feasibility.

The remaining sections are as follows: Section \ref{related_works} covers the literature review regarding on-device training, Online Edge ML and Semi-supervised Edge ML. Section \ref{methodology} describes the proposed methods in more detail and presents an example of use cases. Section \ref{experiments} presents details of datasets used, model selection, experiment design, and evaluation metrics chosen. Section \ref{results} demonstrates results of comparative simulation experiments, and Section \ref{discussion} discusses the implications of results. Finally, Section \ref{conclusion} provides the conclusion of this work.

\section{Related works}\label{related_works}
Edge ML has become a valuable field to study due to its strengths in utilising computation of edge devices, preserving data privacy and security, and being more suitable for real-time applications. We conduct a literature review related to the following perspectives: On-device training involves different aspects of studies, including model optimisation techniques, training algorithms, applications and distributed learning; Online edge ML, which allows training ML models on edge devices incrementally; and the use of semi-supervised learning for edge ML.

\subsection{Edge ML}
One of the challenges for deploying ML models on the edge is the resource constraints of edge IoT devices. To tackle this issue, we need to make the choice between the model performance and the device constraints, which means maintaining a satisfying performance would usually require the devices to have more powerful resource capacity. On the other hand, retaining a satisfying device cost would usually require the model to sacrifice its performance. Hence, investigating the correlation between them would be essential. Research conducted by Gomez-Carmon et al. \cite{ref3} shows that “the trade-off between the computational cost and the model performance could be highly non-linear”, where they have shown that a decline of 3\,\% in model performance could reduce the computational cost up to 80\,\% when the initial data and the early processing were properly used. This reveals the possibility that we could adapt ML models to edge IoT device environments without giving up the power of ML. 

\subsubsection{On-device Training}
An inherited (from centralised ML) implementation for Edge ML follows the steps that: training ML models on a powerful node (e.g., cloud/server) and then deploying them on edge devices, and those devices are used for inference only. On-device training has huge potential due to its strength. Several studies explored the feasibility of training ML models on MCU-based devices directly \cite{ref4, ref5, ref6}. Yazici et al. \cite{ref4} explored the possibility of training classical ML algorithms (Random Forest, SVM and MLP) on devices that have only 1GB of RAM and 1.2 GHz of processing speed. Llisterri Gimenez et al. \cite{ref5} further extended the research of training Deep Neural Networks (DNNs) on similar devices, which is more challenging because there is increased computational cost and more memory required for doing backpropagation. They used a DNN model with one hidden layer (25 hidden neurons and 75 hidden neurons for comparison experiments, respectively) and trained the model for a keyword spotting task. Another work contributed by De Vita et al. \cite{ref6} has focused on training Echo State Networks (ESNs), a type of Recurrent Neural Networks (RNNs), which are generally used for dealing with time series data. 

In 2017, researchers from Google introduced a novel paradigm for on-device training that is based on distributed learning, and they named it Federated Learning (FL) \cite{ref7}. The principles of FL are dispatching, local training and aggregating, which allow ML models to be deployed on edge devices and trained with their local data, then updating the global model by aggregating weights collected from edge devices. The authors in the publication experimented with FL using five different models on four datasets, and the results showed that all models chosen could converge to a good performance on test accuracy and proved that FL is robust enough against unbalanced and non-IID (Independent and Identically Distributed) data. FL provides an effective solution to utilise edge data for training larger models without sacrificing data privacy. 

Further studies have been published investigating and verifying the feasibility of FL on resource-constrained devices \cite{ref8, ref9, ref10}. Das and Brunschwiler \cite{ref8} tested three DL models with federated learning: CNN, LSTM and MLP with the number of model parameters: 47K, 641K and 1.7M, respectively. In \cite{ref9}, similar research to \cite{ref5} was conducted by Llisterri Gimenez et al. where they studied training a shallow DNN with FL and evaluated the effect of hidden neurons for a single hidden layer (5, 10, 20 and 25), and the result is not surprising. By increasing the number of neurons, the model gets quicker to coverage, but this highly depends on the memory capacity of devices. More importantly, they have observed that by increasing the frequency of FL rounds (the process aggregations), the quicker the model’s training loss is going down. Liu et al. \cite{ref10} proposed an FL framework for time series data-related anomaly detection applications in Industrial IoT (IIoT). Their method integrates CNN, LSTM and Attention mechanism. They also employed a gradient compression technique for reducing the number of parameters to send during FL, which enhances communication efficiency against the timeliness concern.

A more straightforward and fundamental way to achieve on-device training is to tackle the models directly, which means if we can make a model lighter and minimise its computational cost and memory usage, then that would be a more general solution to this field. In \cite{ref11}, Choi et al. proposed a co-design method that modifies both software-level and hardware-level design, where the former applies quantisation techniques and layer freezing, and the latter uses a bit-flexible Multiply-and-Accumulate (MAC) array to save memory spaces. Tsukada et al. \cite{ref12} proposed a semi-supervised anomaly detector based on sequential learning for edge IoT devices. Research conducted by Costa et al. \cite{ref13} proposed a lighter version of stochastic gradient descent (L-SGD); they tested SGD and L-SGD on the MNIST dataset and showed that L-SGD can obtain 4.2x faster training speed compared to SGD.

\subsubsection{Online Edge ML}
Edge ML can be achieved by batch learning or incremental learning (online), where the former usually requires a large memory capacity for storing data, and the latter allows data to be fed one by one, which makes it more friendly to resource-constrained devices. 

Sudharsan et al. \cite{ref14} proposed a framework named Edge2Train, which provides an insight into retraining ML models on edge devices. In their work, the acquisition of “Ground Truth” is by observation rather than pre-prepared, which is close to the motivation of our proposed method. In \cite{ref16}, the authors proposed an incremental learning method by transfer learning a pre-trained CNN as the feature extractor component, then use K-Nearest Neighbour (KNN) to make inferences. Ren et al. \cite{ref17} proposed a method called Tiny ML with Online-Learning (TinyOL), which leverages online learning to handle streaming data and the concept drift problem. Research from Sudharsan et al. \cite{ref18} proposed ML-MCU for training (binary and multi-class) classifier models on edge devices, with the streaming data collected and stored on devices directly. They further published another work on using an online learning algorithm called Train++ for IoT devices to achieve self-learning \cite{ref19}. Note that both \cite{ref18, ref19} assumed that the way used to label the collected streaming data is based on the method Edge2Train proposed in \cite{ref14}. Some other research presents various solutions for retraining ML models on edge devices \cite{ref20, ref23, ref24}. Lin et al. \cite{ref20} successfully and practically implemented their methods on a tiny IoT device with 256 KB of SRAM by applying Quantisation-Aware Scaling (QAS) and sparse update. In \cite{ref23}, a micro version of transfer learning (MicroTL) is introduced, which works by applying quantisation to the forward-propagation and de-quantisation during the backward-propagation. They have shown that MicroTL can save 3x less energy consumption compared to the models built in CMSIS-NN \cite{ref24}.

\subsubsection{Semi-supervised Edge ML}
Labelling data or acquiring labelled data has been one of the key challenges in ML, especially for models with supervised paradigms. This problem is more serious with Edge ML because of not only the scarcity of labelled data but also the capability for edge devices to store the entire data set. Semi-supervised learning (SSL) introduced a different way for training ML models, where it trains ML models with only a limited labelled set and then adapts models to learn from a large set of unlabelled data. Several studies \cite{ref25, ref26, ref27, ref28, ref29, ref30, ref31, ref33} have shown how Edge ML can leverage SSL to tackle the shortage of labelled data. 

In \cite{ref25, ref26, ref33}, the authors implemented SSL methods for three specific tasks under edge environments. Xu et al. \cite{ref25} focus on the task of network anomaly detection with fog computing, where they tested a One-Class SVM model with SSL and showed the effectiveness in achieving high accuracy and low latency. Yu et al. \cite{ref26} focused on the task of salient object detection, and they proposed a cloud-edge method that leverages distributed edge nodes (ensemble) to generate pseudo-labels that are used to train a reverse augmented network on the cloud. In \cite{ref33}, Nain et al. proposed EnSeMUp for quality prediction, a common industrial task. EnSeMUp is an SSL framework that integrates Data Augmentation with MixUp strategy, and Ensemble Learning. They tested the method on three real-world manufacturing datasets. 

On the other hand, federated learning inspires researchers due to its characteristics \cite{ref27, ref28, ref29, ref30}. Albaseer et al. \cite{ref27} and Yang et al. \cite{ref28} proposed methods that have a similar training loop: select a subset of IoT devices, train the local model on each device, aggregate updated parameters on the server, and broadcast the global model. In \cite{ref27}, the authors assume that for each IoT device that edge server is connected to, there is labelled data available initially. The work from Tashakori et al. \cite{ref29} aims to provide personalised models for each edge node (user). They proposed semi-supervised personalised FL, where each personalised model for each edge node (users) is an autoencoder model trained by a labelled set on the server, then a small labelled set and a large unlabelled set on the users’ side. The method assumes that there is a set of rich labelled data from different distributions on the server, which can satisfy the creation of the initial autoencoder for each personalised model on the users side. Liu et al. in \cite{ref30} focus on the overfitting problem of semi-supervised FL, which typically happens when training a large model but with limited labelled data provided. They proposed a framework called START that allows progressively training each local model, starting with a very shallow network and then gradually increasing model complexity through iterations. Similar to other works, their method assumes that each client has a certain amount of labelled data for supporting the progressive training. Another distributed SSL method proposed by Chen et al. \cite{ref31} utilises both private labelled data and unlabelled data on each device, where the method generates pseudo-labels by averaging the predictions of neighbouring devices, and also they created a consensus consistency loss function for ensuring pseudo-labels with high consensus are used to train the model. 

To the best of our knowledge, the current algorithms/frameworks do not yet provide an effective way for continuously labelling new unseen data (especially in truly practical scenarios). Several literatures have mentioned this issue (e.g., \cite{ref14}) but did not give a specific implementation for this problem, where some literatures assumed the ground truth label already existed (e.g., \cite{ref13, ref16, ref17, ref23}), or the labels are obtained by less efficient ways, such as expert feedback and manual input (e.g., \cite{ref6, ref8}). Such deficiencies formed the research question we mentioned in Section \ref{intro}.

\section{Methodology}\label{methodology}
\begin{figure}[H]
\centering
\includegraphics[width=1.0\linewidth]{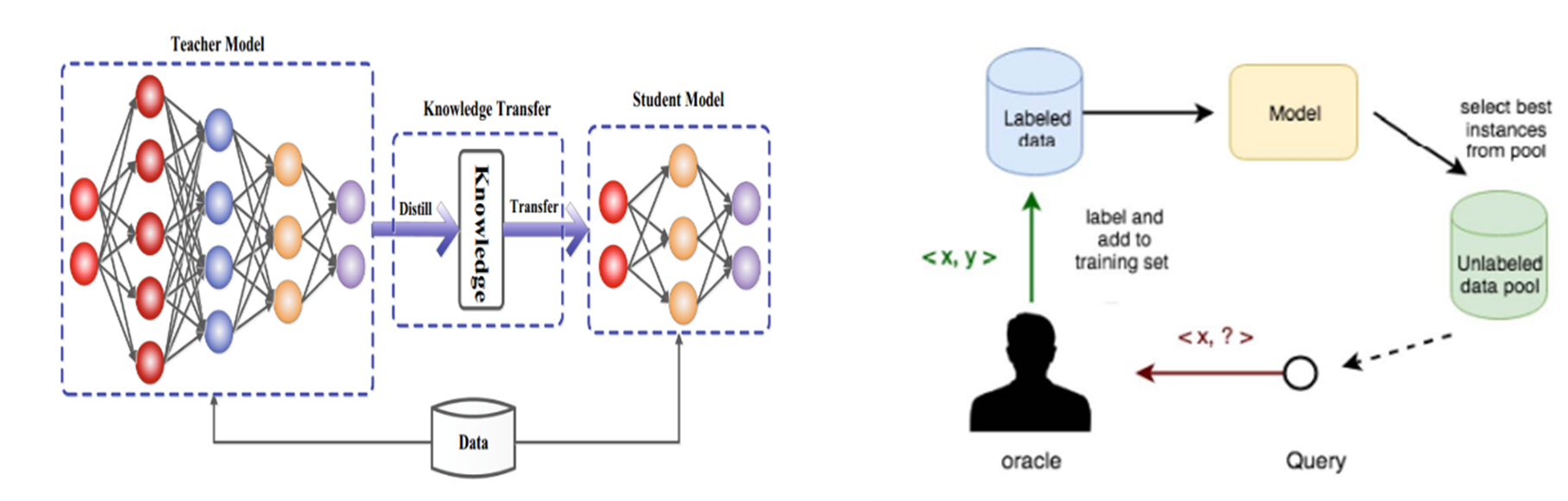}
\caption{Knowledge Distillation (Left) and Active Learning (right) \cite{ref32, ref35}}
\label{fig:KD and AL}
\end{figure}

As described in Section \ref{intro}, the overall system of the proposed method consists of two models, a student model and a teacher model, and it is based on KD and AL. A typical KD process is presented in \hyperref[fig:KD and AL]{Fig.~\ref*{fig:KD and AL}} – Left, which involves using a shared dataset. The teacher model is first pre-trained with such datasets, then the student model is trained by utilising the teacher model’s output (i.e., soft target). The optimal result is that the student model completely mimicked the teacher’s performance (i.e., matching performance) on the same dataset. A typical AL is presented in \hyperref[fig:KD and AL]{Fig.~\ref*{fig:KD and AL}} – Right, the model always selects the most informative data (i.e., the data most likely to minimise the loss) and sends it to the oracle for labelling, then trains the model using the labelled data.

\subsection{Proposed methods}
\begin{figure}
\centering
\includegraphics[width=1.0\linewidth]{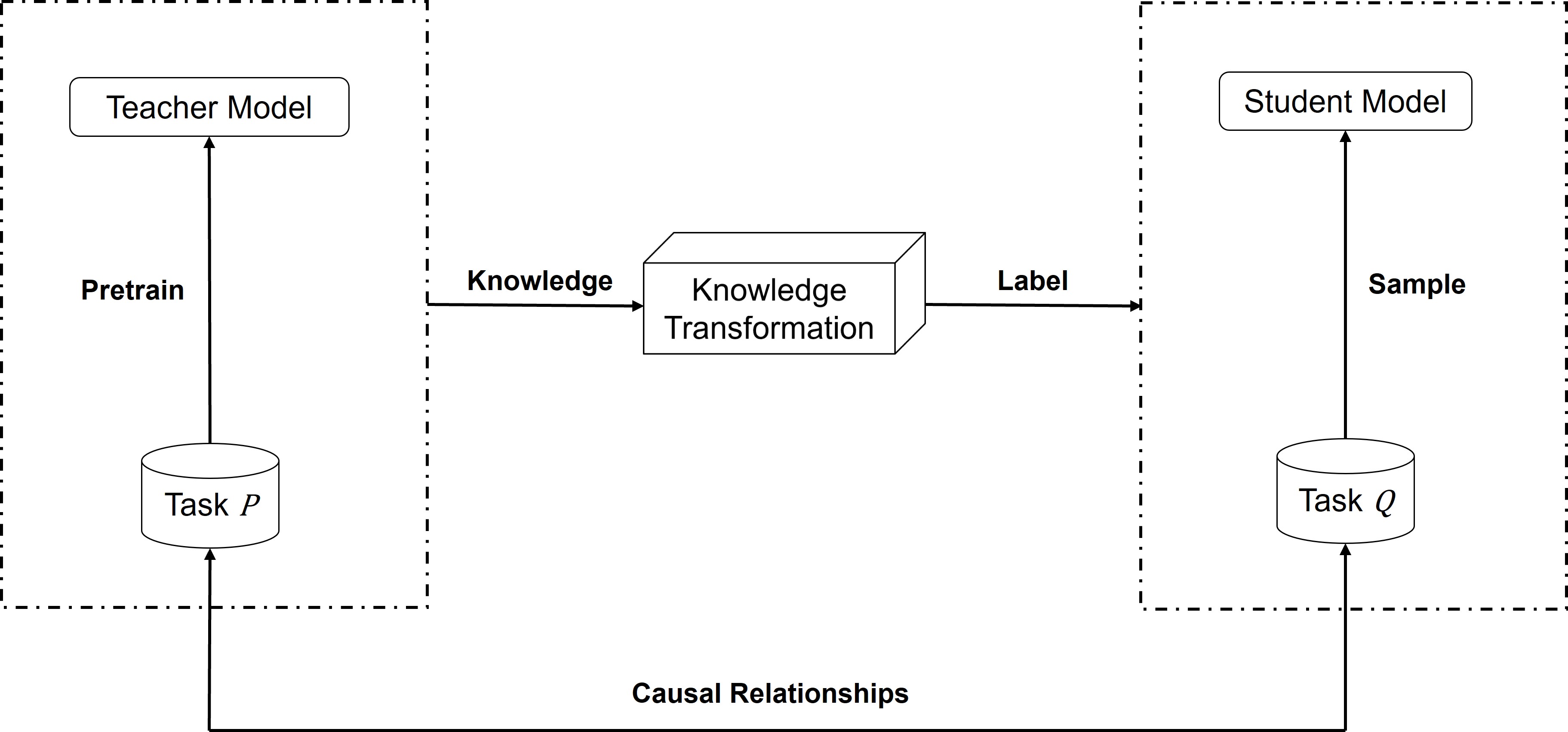}
\caption{KT - Concept Design}
\label{fig:KT_ConceptDesign}
\end{figure}

The proposed method is called Knowledge Transformation (KT). KT asks the teacher model to play the role of the oracle in AL, which directly provides hard labels to support the training of the student model. Unlike traditional KD, the datasets used to train both models can be inconsistent, and the student model is no longer needed (and should not) to mimic the teacher’s behaviour since they are trained on different datasets. The concept design of KT is presented in \hyperref[fig:KT_ConceptDesign]{Fig.~\ref*{fig:KT_ConceptDesign}}.

The teacher model is a model pre-trained on \textit{Task\_P} which we call this model the label generator or the auxiliary model. The student model is the model where the online learning is performed, and it has its own data \textit{Task\_Q}; we call it the online-learning model or the main model. Some key points: (1) The online-learning model needs to be semi-trained with at least $n$ examples, where $n$ is the number of classes (assuming it is dealing with a classification problem), because it needs to at least know how many classes there are. (2) The online-learning model will continue to train after being deployed on edge IoT devices. (3) The label generator has to be a well pretrained, powerful model because it must provide correct label information stably for supporting the online-learning model to train. (4) The label generator model is only responsible for making inferences. The overview of KT is presented in \hyperref[fig:KT_Overview]{Fig.~\ref*{fig:KT_Overview}}.

In practical scenarios, every new data point collected/sampled from \textit{Task\_Q} is naturally unlabelled (completely unseen), and this would require the teacher model to transform its knowledge to generate the corresponding pseudo-labels. Such label generations are achieved via the box "Knowledge Transformation", where the idea is based on the classical propositional logic $P \Rightarrow Q$, which means that if $P$ is true, then $Q$ must also be true.

\subsection{Algorithms}
As shown in Algorithm {\ref{pseudo_code}}, it outlines the training process and the implementation of KT. To better explain them, we have written the pseudo-code in a less abstract way, such as the $S\_classes$ can be seen as an array of strings that contains the classes known by the student model, and the same for $T\_classes$. In this work, the knowledge transformation relies on a simple causal reasoning of $P \Rightarrow Q$, and it can be stored in a dictionary-like data type, which such a data type can be implemented in most programming languages. The $classes\_mapping$ can be seen as a mapping function that says for every class known by the $S\_model$, there exists a corresponding class known by the $T\_model$. Suppose KT is adapted to a co-training scenario where we use two devices to hold the student and teacher models, respectively; then such mapping functions must exist in one of the devices (e.g., stored in the flash memory of the device). The $S\_input$ and $T\_input$ are streaming data captured by both devices, respectively. $T\_\hat{y}$ is the teacher's predicted label ($T\_\hat{y} \in T\_classes$). The while loop (training) ends when certain conditions are met, which we will discuss in Section \ref{discussion}.

\begin{algorithm}
\caption{Training process of KT}
\KwIn{\textit{T\_model, S\_model}, Data stream for the teacher and student model, respectively: $T\_input = \{x_1^{(T)}, x_2^{(T)}, x_3^{(T)}, \dots \}$, $S\_input = \{x_1^{(S)}, x_2^{(S)}, x_3^{(S)}, \dots \}$}

$T\_classes \gets \{c_0^{(T)}, c_1^{(T)}, c_2^{(T)}, \dots, c_n^{(T)}\}$\;

$S\_classes \gets \{c_0^{(S)}, c_1^{(S)}, c_2^{(S)}, \dots, c_n^{(S)}\}$\;

$classes\_mapping \gets \{c_0^{(T)}:c_0^{(S)}, c_1^{(T)}:c_1^{(S)}, c_2^{(T)}:c_2^{(S)}, \dots , c_n^{(T)}:c_n^{(S)} \}$\;

\While{$x_i^{(S)}$ \textbf{and} $x_i^{(T)}$}{
    $T\_logits \gets T\_model(x_i^{(T)})$\;
    $S\_logits \gets S\_model(x_i^{(S)})$\;
    $T\_\hat{y} \gets argmax(Softmax(T\_logits))$\;
    $pseudo\_label \gets classes\_mapping(T\_\hat{y})$\;
    $S\_loss \gets L(S\_logits, pseudo\_label)$\;
    $ S\_model.backward()$
}
\label{pseudo_code}
\end{algorithm}

\subsection{Formulation and use cases}
KT supposes a scenario in which there exist at least two events/tasks, $P$ and $Q$, and for each task, the future data points are sampled from certain (unknown) distributions. The distribution of both tasks can be independent from each other. 

\begin{equation}
    \forall x_i^{(P)} \sim D_P, \forall x_i^{(Q)} \sim D_Q
\end{equation}
\begin{equation}
    D_P \perp\!\!\!\perp D_Q
\end{equation}

Suppose $Q$ is a task of classifying items that keep showing up (the "cause"), and $P$ is the task of observing humans' reactions based on the item appearing in $Q$ (the "effect"). Both tasks can be treated as image classification tasks, and now we use a pre-trained teacher model $f_\theta^{(T)}$ to handle the task $P$ and an online-learning student model $f_\theta^{(S)}$ to constantly learn the task $Q$. As described in Eqs.~\eqref{eqs:teacher_model} and \eqref{eqs:student_model}, the $\eta$ is the learning rate of the student model, then $y_i^{(S)}$ is the true label for the $i^{th}$ data point. The generation of $y_i^{(S)}$ relies on the knowledge transformation function in Eqs.~\eqref{eqs:KT}, and its definition is described in Eqs.~\eqref{eqs:KT_definition}, where $\mathcal{Y^{(S)}}$ and $\mathcal{Y^{(T)}}$ represent the set of labels known by the student model and the teacher model, respectively. Once we start training the student model, $\hat{y}^{(T)}$ will substitute $y^{(T)}$. 

\begin{equation}\label{eqs:teacher_model}
    \hat{y}_i^{(T)} = f_\theta^{(T)}(x_i^{(P)}) 
\end{equation}

\begin{equation}\label{eqs:student_model}
    \theta^{(S)} = \theta^{(S)} - \eta \cdot \nabla_{\theta^{(S)}} L(f_{\theta}^{(S)}(x_i^{(Q)}), y_i^{(S)}) 
\end{equation}

\begin{equation}\label{eqs:KT}
    KT: f(\hat{y}^{(T}) \to y^{(S)}
\end{equation}

\begin{equation}\label{eqs:KT_definition}
    \forall y^{(S)} \in \mathcal{Y^{(S)}}, \exists! y^{(T)} \in \mathcal{Y^{(T)}} (y^{(T)} \Rightarrow y^{(S)})
\end{equation}

The definition of KT is saying that for every input $x_i^{(Q)}$, there exists a corresponding output $y_i^{(T)}$, and we can infer the true label $y_i^{(S)}$ based on the causal relationship we assumed. Hence, the inferred result can be seen as the ground truth label of $x_i^{(Q)}$.

\begin{figure}[H]
\centering
\includegraphics[width=1.0\linewidth]{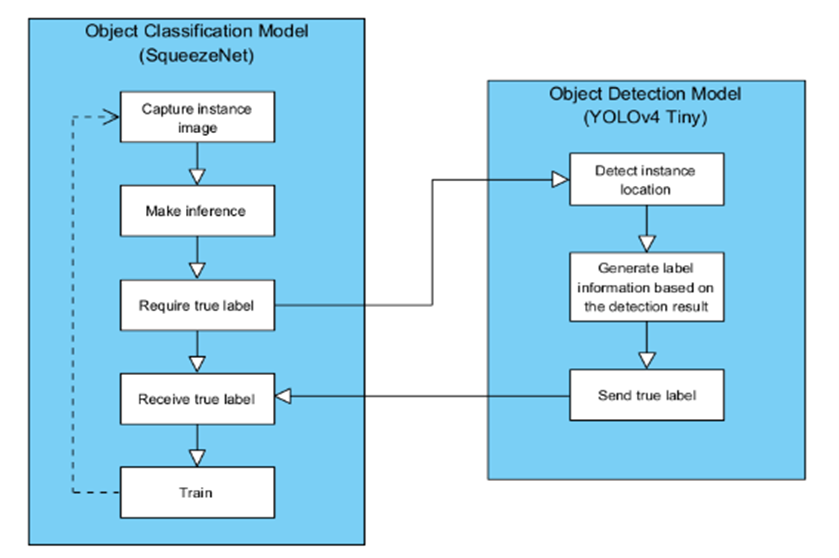}
\caption{KT - Use Case Illustration}
\label{fig:KT_Usecase}
\end{figure}

An instance of use cases based on the proposed method is illustrated in \hyperref[fig:KT_Usecase]{Fig.~\ref*{fig:KT_Usecase}}. Suppose the scenario is in a production line, where workers inspect products and take some “action” if they observe that there are products of “bad” quality (e.g., “move” such products off the line). The task of the label generator, which is the object detection model on the right is to detect such human reactions and then generate corresponding labels of products (e.g., good or bad). Our goal is to train the object classification model (on the left) to automatically classify the quality of products, and this is achieved by continuously training this model, where the ground truth label can be obtained by the causal reasoning that $product\_moved \Rightarrow bad\_product$. This is a basic use case of KT because both models are simply two binary classification models, but this outlines the core idea of the proposed method.

\section{Experiments}\label{experiments}

\subsection{Datasets}
\begin{figure}[H]
\centering
\includegraphics[width=1.0\linewidth]{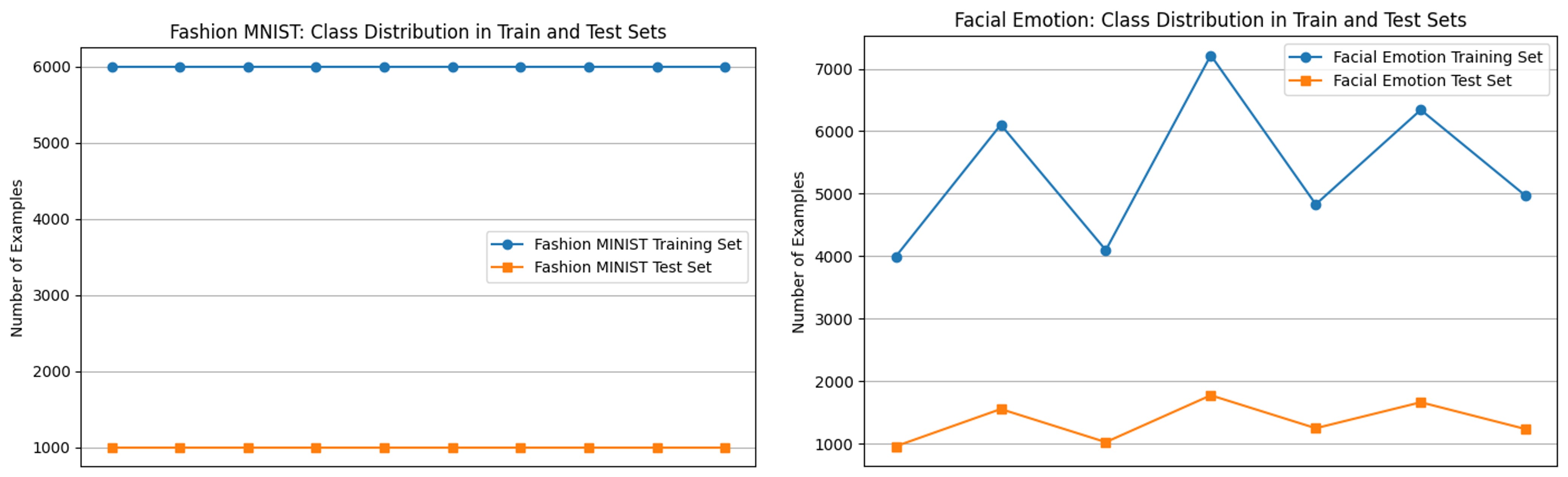}
\caption{KT - Class distribution of Fashion MNIST (left) and Facial Emotion (right)}
\label{fig:class distributions of two datasets}
\end{figure}

\begin{figure}[H]
\centering
\includegraphics[width=1.0\linewidth]{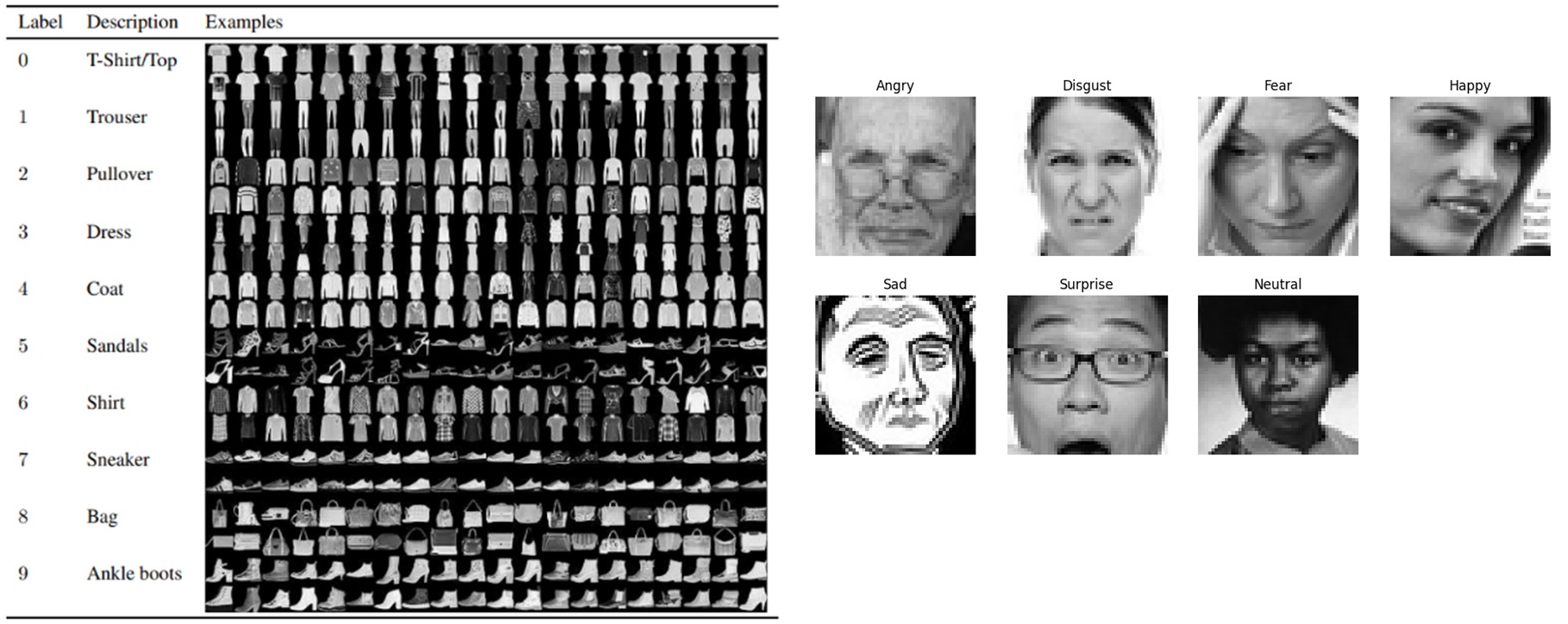}
\caption{Fashion MNIST data (left)\cite{ref36}, Facial Emotion data (right)}
\label{fig:example data from both datasets}
\end{figure}

To conduct the simulation experiments, we use two datasets for the online-learning model and the label generator model respectively, Fashion MNIST \cite{ref36} and Facial Emotion \cite{ref37}. Fashion MNIST is a benchmark dataset for image classification models in ML, and it is embedded in mainstream DL frameworks, such as PyTorch \cite{ref21}, Scikit-learn \cite{ref15} and Keras \cite{ref22}. The latter is a new dataset available on Kaggle \cite{ref38}, which contains images of human facial emotions. Images of Fashion MNIST are fixed with the size of 28*28*1 and there are 10 classes in the dataset: T-Shirt/Top, Trouser, Pullover, Dress, Coat, Sandals, Shirt, Sneaker, Bag, and Ankle boots. Images of the Facial Emotion dataset are fixed with the size of 48*48*1 (we modified this to 40*40*3 during simulation experiments), and there are 7 classes in the dataset: Angry, Disgust, Fear, Happy, Sad, Surprise, and Neutral. Note that the third channel of image size refers to the colour of images, hence for both datasets they are all greyscale images originally. The distribution of the training set and test set for both datasets is
presented in \hyperref[fig:class distributions of two datasets]{Fig.~\ref*{fig:class distributions of two datasets}}. More detail is given below. Fashion MNIST's training set: 6,000 examples per each class, 60,000 in total; Fashion MNIST's test set: 1,000 examples per each class, 10,000 in total. The class distribution of the Facial Emotion dataset is relatively imbalanced compared to the Fashion MNIST, where its training set is: [3995, 6104, 4097, 7215, 4830, 6342, 4965] and its test set is: [958, 1554, 1024, 1774, 1247, 1662, 1233]. \hyperref[fig:example data from both datasets]{Fig.~\ref*{fig:example data from both datasets}} presents example images from both datasets.

\subsection{Model selection}
\label{subsec:model selection}
Intuitively, we may realise that classifying human emotions is a much harder task than the Fashion MNIST task, which means the model that tackles the former would have to be complex enough (i.e., more parameters). This study is not about finding certain architectures that are best for certain tasks. Instead of that, it cares about whether the student model can indeed improve itself by using the knowledge transformed from the teacher. Hence, the ablation experiments are based on the same network architecture that is used for both the teacher model and the student model.

SqueezeNet, an architecture published in 2016, in the paper "SqueezeNet: AlexNet-Level Accuracy with 50x Fewer Parameters and $<$ 0.5MB Model Size" \cite{ref39}. The authors proved that SqueezeNet can obtain a Top-1 accuracy of 57.5\,\% on ImageNet \cite{ref40} with only 0.47 MB of model parameters after compressions, which compared to AlexNet \cite{ref41}, which requires 240 MB of model parameters to achieve the same level of performance. 

\begin{figure}
\centering
\includegraphics[width=1.0\linewidth]{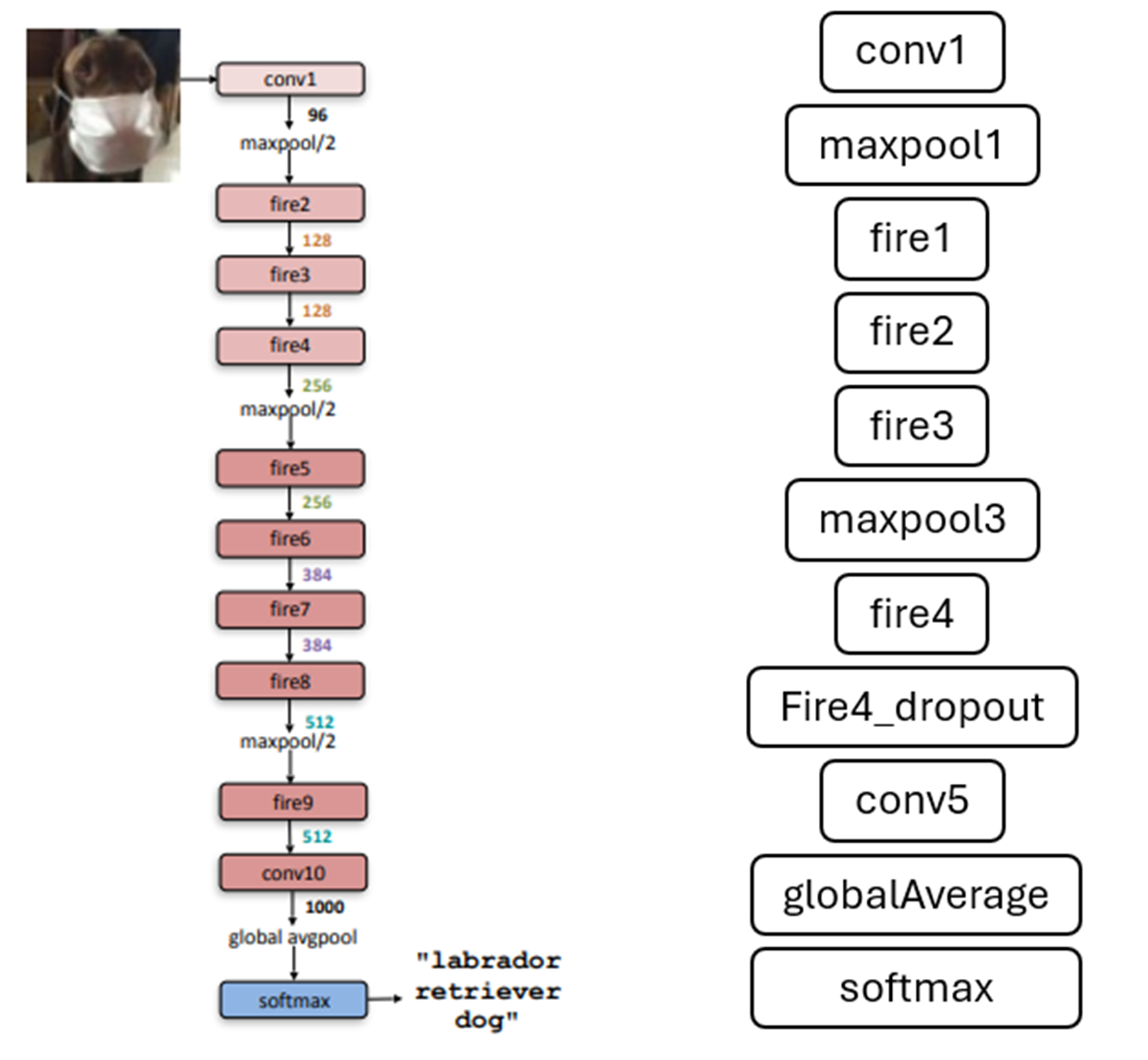}
\caption{Original SqueezeNet (left) \cite{ref39}, Simplified SqueezeNet (right)}
\label{fig:simplified SqueezeNet}
\end{figure}

We adapt SqueezeNet to the two tasks in our research (Fashion MNIST and Facial Emotion). The original SqueezeNet was built for a more complex task (ImageNet-1k) and the input size is different (it used for 224*224*3), therefore, we used a simplified version of SqueezeNet that built with fewer fire modules (only 4) and reduced the number of feature maps of convolutional layers. The architecture of the modified SqueezeNet is presented in \hyperref[fig:simplified SqueezeNet]{Fig.~\ref*{fig:simplified SqueezeNet}}. The specific configuration of the simplified SqueezeNet is given in \hyperref[table:configuration of SqueezeNet]{Table.~\ref*{table:configuration of SqueezeNet}}.

\begin{table}[H]
\caption{Parameters of Modules}
\centering
\begin{tabular}{@{}ll@{}}
\toprule
Module & Configuration \\
\midrule
conv1 & feature\_map = 16; kernel\_size = (3,3), stride = (2,2)\\
maxpool1 & pool\_size = (3,3); strides = (2,2)\\ 
fire1 & squeezeLayer\_fm = 4; expandLayer\_fm = 16\\ 
fire2 & squeezeLayer\_fm = 4; expandLayer\_fm = 16\\ 
fire3 & squeezeLayer\_fm = 8; expandLayer\_fm = 32\\ 
maxpool3 & pool\_size = (3,3); strides = (2,2)\\ 
fire4 & squeezeLayer\_fm = 8; expandLayer\_fm = 32\\ 
fire4\_dropout & dropout\_rate = 0.5\\ 
conv5 & feature\_map = number of classes; kernel\_size = (1,1) \\
\bottomrule
\end{tabular}
\label{table:configuration of SqueezeNet}
\end{table}

Note that we did not change the design of fire modules, which is the principle of SqueezeNet. Fire modules in the modified version still follow the original design that one squeezing layer followed by two expanding layers, and the hyperparameter kernel\_size for these components remained unchanged: (1,1) for squeezing and (3,3) for expanding. The variables “squeezeLayer\_fm” and “expandLayer\_fm” in the configuration refer to the number of feature maps used for creating the squeezing layer and the expanding layer, respectively. All padding schemes applied to convolutional layers remain the original setting. There are two major changes: for all convolutional layers, we changed the kernel initialisation method (weight initialising) to “he\_uniform” \cite{ref42} and the activation function to “mish” \cite{ref43}. The reason why we made these two changes is because we have attempted several configurations, and the current setting obtained better performance in accuracy. The following terms are used interchangeably throughout the remaining content. Student model: main model or online-learning (OL) model; Teacher model: auxiliary model or label generator model.

\subsection{Experiment design}
In real scenarios, the facial emotion dataset will be the one to pretrain the label generator model because it is related to humans (i.e., generate labels based on facial emotions), and the Fashion MNIST will be the one that the main model deals with because it is related to items (i.e., classify items). To simulate the online learning process, we need to split the Fashion MNIST into a semi-training set and an online-learning set, and for the Facial Emotion dataset, we need to split it into a pre-training set and an online-learning set. For each dataset, the original training set and test set can be merged together. 

There are three key things:
\begin{enumerate}
    \item Both models should have the same number of classes, because based on the causal relationship we defined in the formulation, every item has exactly one corresponding human reaction; hence, if there are 10 classes used in the Fashion MNIST, then we need to have 10 facial emotions to correspond to each of them. Due to the inconsistency of class numbers (10 in Fashion MNIST, 7 in Facial Emotion), our simulation experiments are conducted on the basis that only 7 classes are used.
    \item For both online-learning sets constructed, the data has to be placed in the identical order. It means that if the order of the online-learning set for the main model is: $\{0,1,3,5,1,2,4 \dots, \dots \}$ where the number represents the target label of each data point, then the order of the online-learning set for the auxiliary model needs to be the same as that. This is crucially important for simulations (but not for reality) because if we have an incorrect facial emotion corresponding to an item, then this will generate incorrect label information even when the label generator has made a correct prediction. In reality, this should not be the case, because for every item there must be a corresponding human reaction, and it is very likely implying the item’s class. This is why it is so called “cause and effect.”
    \item Lastly, for both online-learning sets constructed, they should contain samples only and no target labels are provided. Because target labels are used for training, where the label generator has been designed to make inferences only and the main model uses the “ground truth label” transformed from the label generator.
\end{enumerate}
For the online-learning model, we assume that such models are initially trained with only one example per class, so the semi-training set for the online-learning model will be 7 examples in total (assume 7 classes are used). The size of the pretraining set for the label generator model differs depending on the size chosen for the online-learning set, which can have two cases: the online-learning set is balanced or imbalanced (the latter is more usual). We have conducted experiments regarding these two situations.

\begin{table}[H]
\caption{Training Settings}
\centering
\begin{tabular}{@{}ll@{}ll@{}}
\toprule
Setting & Online-learning model & Label generator model \\
\midrule
Loss function & SCC & SCC\\ 
Optimiser & Adam & Adam\\ 
Metrics & Accuracy & Accuracy\\ 
Epochs & 10 & 100 \\ 
Batch size & 1 & 128 \\ 
Validation ratio & None & 20\,\% \\ 
ModelCheckPoint & monitor = ‘loss’ & monitor = ‘val\_loss’\\ 
\bottomrule
\end{tabular}
\label{table:training settings}
\end{table}

The training settings for both models in stage one (train the label generator model with the pre-training set and train the online-learning model with the semi-training set) are presented in \hyperref[table:training settings]{Table.~\ref*{table:training settings}}, SCC refers to Sparse Categorical CrossEntropy. The epochs for the initial training of the OL model do not necessarily need to be a large number since there are not that much data to train. Whereas, for the label generator model, we need to make sure it has been well pretrained, therefore, training it with more epochs is very essential. The semi-training set does not have enough data to split a validation set, hence the validation set is only used for the label generator model. And therefore, the monitor used to store the checkpoint is different. Note: All experiments are simulated on a Windows operating system with a CPU of 4 cores and a lock speed of 2.1 GHz, and 8 GB of SRAM (6.94 GB usable during experiments). The framework used to build and train the model is Keras with version 3.4.1.

\subsection{Evaluation metrics}
The evaluations for both models are slightly different. For the label generator model, we want to assess the model performance on its online-learning dataset (which is actually the test set for it), then for the OL model we want to assess its expected maximum performance (when it is trained with correct labels) and its actual performance (when it is trained with labels provided by the label generator model). 

The most straightforward metric to evaluate a classification model is accuracy, which it simply sums all true class predictions (True Positive and True Negative) then divides by the sum of all predictions including false class predictions (False Positive and False Negative).

\begin{equation}\label{eqs:accuracy}
    Accuracy = \frac{TP + TN}{TP + TN + FP + FN}
\end{equation}

Since the label generator and the OL model can be multiclass classification models (when the number of classes to use exceeds 2), then simply using accuracy as the only metric to evaluate the model performance is not suitable, especially when the data is imbalanced. Therefore, we also employed the metrics of precision, recall and F1 score. The precision tells us how many predictions that the model made are actually the class of the data points (True Positive), for example, the label generator model predicts 12 facial emotions as happy out of 20 faces, where 10 of those 12 predictions are actually “happy face”, so the model predicted 10 True Positive and 2 False Positive for the class “happy”, and the model obtained a precision of 83.3\,\% on the class “happy”.

\begin{equation}\label{eqs:precision}
    Precision = \frac{TP}{TP+FP}
\end{equation}

The recall indicates the number of True Positive have been observed by the model. Using the same example above, where there were actually 18 happy faces in those 20 faces, and the model missed 8 happy faces, which means the model obtained a recall of 55.6\,\% on the class “happy”.

\begin{equation}\label{eqs:recall}
    Recall = \frac{TP}{TP+FN}
\end{equation}

The F1-score provides a more comprehensive interpretation by balancing the precision and recall, the score ranges from 0 to 1, and the score of 1 indicates a perfect model (on that dataset) that performs well on both precision and recall. 
\begin{equation}\label{eqs:f1_score}
    F1 = \frac{2(Precision * Recall)}{Precision + Recall}
\end{equation}

In the experiments, we evaluated the performance of both models using these metrics, and we scaled the number of classes to use,  starting from a basic binary classification to a multiclass classification of 7 classes. We also evaluated the F1 score on each class. 

\section{Results}\label{results}

\subsection{Weak teacher model}
In this part, the teacher model is assigned to handle the Facial Emotion data and the student model will be trained to classify Fashion MNIST items.  

\begin{table}
\caption{Performance of the label generator model (Facial Emotion) from a binary classification to a 7-class classification}
\begin{tabular}{@{}cccccccc@{}}
\toprule
\textbf{Classes} & \textbf{Acc (\%)} & \multicolumn{2}{c}{\textbf{Precision}} & \multicolumn{2}{c}{\textbf{Recall}} & \multicolumn{2}{c}{\textbf{F1 Score}} \\
&  & \textbf{Micro} & \textbf{Macro} & \textbf{Micro} & \textbf{Macro} & \textbf{Micro} & \textbf{Macro}    \\ 
\midrule
\multicolumn{8}{c}{\textbf{Balanced OL set}} \\
\midrule
2 & 90.4 & 0.904 & 0.905 & 0.904 & 0.904 & 0.904 & 0.904 \\
3 & 71.1 & 0.711 & 0.706 & 0.711 & 0.711 & 0.711 & 0.706 \\
4 & 66.0 & 0.660 & 0.662 & 0.660 & 0.660 & 0.660 & 0.647 \\
5 & 55.1 & 0.551 & 0.547 & 0.551 & 0.551 & 0.551 & 0.535 \\
6 & 54.0 & 0.540 & 0.525 & 0.540 & 0.540 & 0.540 & 0.521 \\
7 & 52.2 & 0.522 & 0.508 & 0.522 & 0.522 & 0.522 & 0.502 \\
\midrule
\multicolumn{8}{c}{\textbf{Imbalanced OL set}} \\
\midrule
2 & 90.4 & 0.904 & 0.904 & 0.904 & 0.904 & 0.904 & 0.904 \\
3 & 73.0 & 0.730 & 0.694 & 0.730 & 0.697 & 0.730 & 0.694 \\
4 & 69.1 & 0.691 & 0.650 & 0.691 & 0.643 & 0.691 & 0.635 \\
5 & 57.7 & 0.577 & 0.531 & 0.577 & 0.438 & 0.577 & 0.530 \\
6 & 59.7 & 0.597 & 0.559 & 0.597 & 0.560 & 0.597 & 0.557 \\
7 & 55.4 & 0.554 & 0.520 & 0.554 & 0.521 & 0.554 & 0.517 \\
\bottomrule
\end{tabular}
\label{table:label generator performance on OL set, after pretraining}
\end{table}

To simulate the scenario when a balanced OL set is sampled. For the Facial Emotion dataset, we take 1,500 examples (samples only) for each class to construct the online-learning set, and then the rest of the examples are used for the pretraining phase of the label generator model. For the Fashion MNIST dataset, we do the same thing for constructing its online-learning set, and then we give 1 example per class to construct the semi-training set for the initial training phase of the online-learning model. 

To simulate the second scenario (imbalanced OL dataset). For the Facial Emotion dataset, we manually select the amount of examples per each class to construct the pretraining set: [4000, 6500, 4500, 7500, 5000, 7000, 5000], which leaves a less balanced dataset when constructing the OL set: [953, 1158, 621, 1489, 1077, 1004, 1198]. For the Fashion MNIST dataset, 1 example per each class (semi-training set), then its OL set is the same as the label generator model uses (in quantity).

For the label generator model, \hyperref[table:label generator performance on OL set, after pretraining]{Table.~\ref*{table:label generator performance on OL set, after pretraining}} shows its performance on its own OL set, from a basic binary classification to a 7-classes classification. We can see that the model accuracy quickly drops down as the number of classes used increases, and even with the minimum number of classes used (2 classes), this model can only reach 90.4\% accuracy and 0.904 macro f1 score. \hyperref[table:label generator performance on each class, after pretraining]{Table.~\ref*{table:label generator performance on each class, after pretraining}} shows the (macro) F1 score of each class, which is more important in our proposed method because the label generator model needs to guarantee to have a rather acceptable level of performance across all classes. If the label generator performs extremely poorly on certain classes, then that will seriously influence the main model to train those classes.

\begin{table}[H]
\caption{Macro F1-score the label generator (Facial Emotion) obtains on each class}
\begin{tabular}{@{}cccccccc@{}}
\toprule
\textbf{Classes} & \textbf{Angry} & \textbf{Disgust} & \textbf{Fear} & \textbf{Happy} & \textbf{Sad} & \textbf{Surprise} & \textbf{Neutral} \\ 
\midrule
\multicolumn{8}{c}{\textbf{Balanced OL set}} \\
\midrule
2 & 0.906 & 0.902 &  &  &  &  &  \\ 
3 & 0.636 & 0.883& 0.598 &  &  &  &  \\
4 & 0.552 & 0.893 & 0.443 & 0.7 &  &  &  \\ 
5 & 0.406 & 0.861 & 0.327 & 0.637 & 0.441 &  & \\ 
6 & 0.315 & 0.834 & 0.244 & 0.640 & 0.408 & 0.684 &  \\ 
7 & 0.353 & 0.822 & 0.209 & 0.610 & 0.375 & 0.690 & 0.456 \\ 
\midrule
\multicolumn{8}{c}{\textbf{Imbalanced OL set}} \\
\midrule
2 & 0.899 & 0.909 &  &  &  &  &  \\ 
3 & 0.645 & 0.904 & 0.532 &  &  &  &  \\ 
4 & 0.461 & 0.877 & 0.426 & 0.777 &  &  &  \\ 
5 & 0.401 & 0.845 & 0.323 & 0.710 & 0.374 &  & \\ 
6 & 0.396 & 0.839 & 0.256 & 0.711 & 0.442 & 0.70 &  \\ 
7 & 0.346 & 0.842 & 0.205 & 0.669 & 0.378 & 0.693 & 0.488 \\ 
\bottomrule
\end{tabular}
\label{table:label generator performance on each class, after pretraining}
\end{table}

\begin{figure}[H]
\centering
\includegraphics[width=1.0\linewidth]{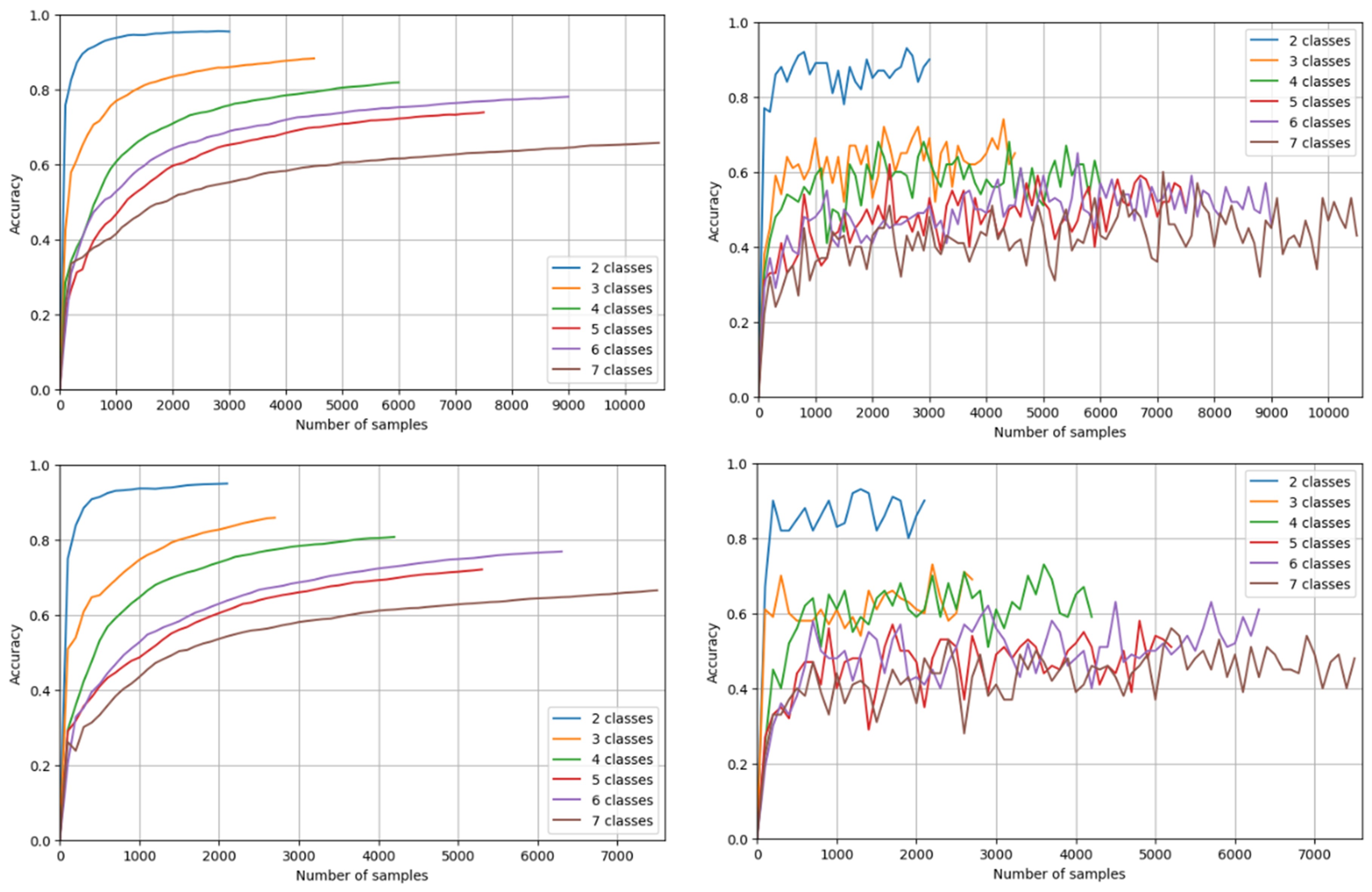}
\caption{Training process of OL model (Expected vs Actual)}
\label{fig:traing process of OL model - 1}
\end{figure}

\begin{figure}[H]
\centering
\includegraphics[width=1.0\linewidth]{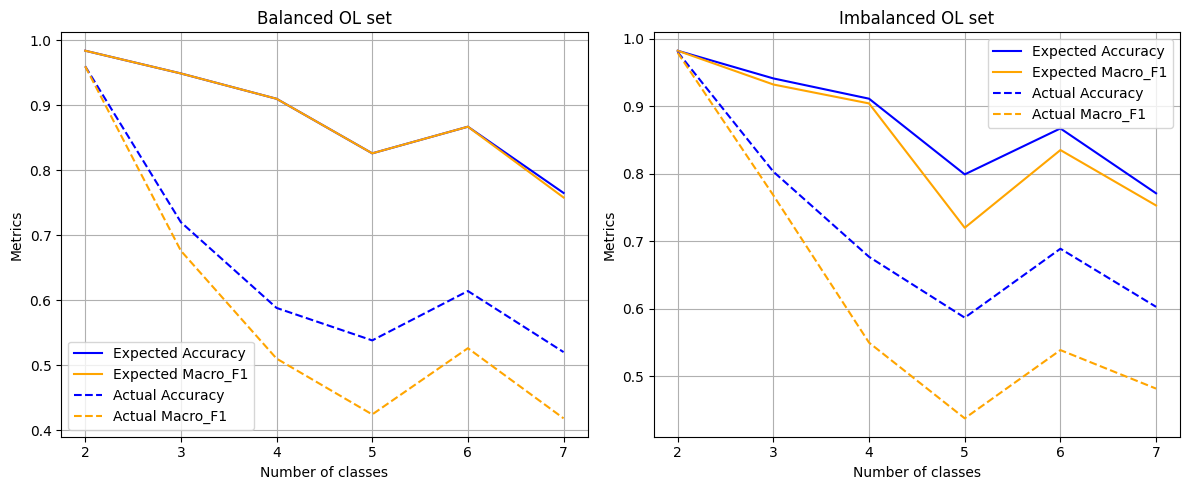}
\caption{Final performance of OL model (Expected vs Actual)}
\label{fig:final performance of OL model - 1}
\end{figure}

For the online-learning model, we recorded the change of model performance (for every 100 samples trained) throughout the training process. In \hyperref[fig:traing process of OL model - 1]{Fig.~\ref*{fig:traing process of OL model - 1}}, the graphs on the left indicate that with correct labels provided, the OL model should have such trends when different numbers of classes are used (expected convergence). The graphs on the right show the actual training process of the OL model when it relies on the labels provided by the label generator model. Furthermore, we evaluated the final performance of the OL model on its OL sets after the stage two (online-learning) is completed. In \hyperref[fig:final performance of OL model - 1]{Fig.~\ref*{fig:final performance of OL model - 1}}, the solid lines represent when the supervision signal is provided by correct labels, the expected performance (accuracy and macro F1 score) the OL model should achieve, and the dashed lines represent the actual performance when the OL model is supervised by the generated pseudo-labels. \hyperref[table:OL model performance on each class - balanced OL set]{Tables.~\ref*{table:OL model performance on each class - balanced OL set}} and \hyperref[table:OL model performance on each class - imbalanced OL set]{\ref*{table:OL model performance on each class - imbalanced OL set}} show the OL model's performance on each class, which explains the underlying reason why the actual final performance is far different from the expected one.

\begin{table}[h!]
\caption{Balanced OL set: Macro F1-score the OL model obtains on each class}
\begin{tabular}{@{}cccccccccccccccc@{}}
\toprule
\textbf{Classes} & \textbf{T-shirt} & \textbf{Trouser} & \textbf{Pullover} & \textbf{Dress} & \textbf{Coat} & \textbf{Sandal} & \textbf{Shirt} \\ 
\midrule
\multicolumn{8}{c}{\textbf{Expected}} \\
\midrule
2 & 0.962 & 0.958 &  &  &  &  &  \\ 
3 & 0.931 & 0.962 & 0.953 &  &  &  &  \\ 
4 & 0.879 & 0.961 & 0.925 & 0.875 &  &  &  \\ 
5 & 0.869 & 0.922 & 0.743 & 0.823 & 0.776 &  & \\ 
6 & 0.896 & 0.971 & 0.723 & 0.854 & 0.772 & 0.986 &  \\ 
7 & 0.747 & 0.958 & 0.684 & 0.790 & 0.670 & 0.986 & 0.470 \\ 
\midrule
\multicolumn{8}{c}{\textbf{Actual}} \\
\midrule
2 & 0.962 & 0.958 &  &  &  &  &  \\ 
3 & 0.703 & 0.975 & 0.350 &  &  &  &  \\ 
4 & 0.388 & 0.969 & 0 & 0.683 &  &  &  \\ 
5 & 0 & 0.960 & 0 & 0.655 & 0.507 &  & \\ 
6 & 0 & 0.951 & 0 & 0.748 & 0.475 & 0.980 &  \\ 
7 & 0 & 0.943 & 0 & 0.605 & 0.406 & 0.975 & 0 \\ 
\bottomrule
\end{tabular}
\label{table:OL model performance on each class - balanced OL set}
\end{table}

\begin{table}[H]
\caption{Imbalanced OL set: Macro F1-score the OL model obtains on each class}
\begin{tabular}{@{}cccccccccccccccc@{}}
\toprule
\textbf{Classes} & \textbf{T-shirt} & \textbf{Trouser} & \textbf{Pullover} & \textbf{Dress} & \textbf{Coat} & \textbf{Sandal} & \textbf{Shirt} \\ 
\midrule
\multicolumn{8}{c}{\textbf{Expected}} \\
\midrule
2 & 0.980 & 0.984 &  &  &  &  &  \\ 
3 & 0.925 & 0.979 & 0.893 &  &  &  &  \\ 
4 & 0.879 & 0.973 & 0.852 & 0.911 &  &  &  \\ 
5 & 0.852 & 0.962 & 0.232 & 0.841 & 0.714 &  & \\ 
6 & 0.858 & 0.964 & 0.529 & 0.874 & 0.798 & 0.990 &  \\ 
7 & 0.753 & 0.941 & 0.555 & 0.852 & 0.708 & 0.975 & 0.485 \\ 
\midrule
\multicolumn{8}{c}{\textbf{Actual}} \\
\midrule
2 & 0.977 & 0.982 &  &  &  &  &  \\ \
3 & 0.721 & 0.977 & 0.607 &  &  &  &  \\ 
4 & 0.008 & 0.921 & 0.527 & 0.742 &  &  &  \\ 
5 & 0.391 & 0.944 & 0.085 & 0.771 & 0 &  & \\ 
6 & 0 & 0.904 & 0 & 0.791 & 0.567 & 0.974 &  \\ 
7 & 0 & 0.940 & 0 & 0.712 & 0.515 & 0.981 & 0.223 \\ 
\bottomrule
\end{tabular}
\label{table:OL model performance on each class - imbalanced OL set}
\end{table}

\subsection{Stronger teacher model}
From the experiment above, we attempt to simulate the scenario that is more lifelike to the real scenario the proposed method will be deployed in, where we used the Fashion MNIST for the main model and the Facial Emotion for the auxiliary model. However, as mentioned in \ref{subsec:model selection}, for the purpose of ablation experiments, we use the same network architecture for both models. This setup allows us to isolate and evaluate the effectiveness of the proposed method, rather than differences arising from the inherent learning capabilities of the student model. The architecture has only 8479 parameters in total (33.12 KB of memory) for the auxiliary model, which is not complex enough to handle the Facial Emotion dataset. A not powerful enough label generator model cannot stably provide trustworthy labels, which may result in the actual training process and final performance of the main model suffering from serious offsets (\hyperref[fig:traing process of OL model - 1]{Figures.~\ref*{fig:traing process of OL model - 1}} and \hyperref[fig:final performance of OL model - 1]{\ref*{fig:final performance of OL model - 1}}) compared to its expected performance.  

In this part, we swap the datasets for both models, so now the label generator model is assigned to handle the Fashion MNIST, and the main model will be learning to classify human emotions. Then, we conduct the experiments based on the setting of a balanced OL set used. In detail, for the Fashion MNIST, we take 3,000 examples per class to construct the pre-training set (21,000 examples in total), then 4,000 examples per each class to construct the online-learning set (28,000 examples in total).

The label generator model now is handling a much easier task, and we can see the difference in \hyperref[table:label generator performance on OL set, after pretraining - stronger teacher]{Table.~\ref*{table:label generator performance on OL set, after pretraining - stronger teacher}}. It shows how the label generator model (with the same architecture) performs on its OL set over a different task (compared to \hyperref[table:label generator performance on OL set, after pretraining]{Table.~\ref*{table:label generator performance on OL set, after pretraining}}). In the meantime, if we look at the F1 score of the label generator model on each class, in \hyperref[table:label generator performance on each class, after pretraining - stronger teacher]{Table.~\ref*{table:label generator performance on each class, after pretraining - stronger teacher}}, it now achieves a rather acceptable level of performance.

\begin{table}[h!]
\caption{Performance of the label generator model (Fashion MNIST) from a binary classification to a 7-class classification}
\begin{tabular}{c c cc cc cc}
\toprule
\textbf{Classes} & \textbf{Acc (\%)} & \multicolumn{2}{c}{\textbf{Precision}} & \multicolumn{2}{c}{\textbf{Recall}} & \multicolumn{2}{c}{\textbf{F1 Score}} \\ 
 & & \textbf{Micro} & \textbf{Macro}       & \textbf{Micro} & \textbf{Macro}    & \textbf{Micro} & \textbf{Macro}     \\ 
\midrule
2 & 98.4 & 0.984 & 0.984 & 0.984 & 0.984 & 0.984 & 0.984 \\ 
3 & 96.4 & 0.964 & 0.964 & 0.964 & 0.964 & 0.964 & 0.963 \\ 
4 & 93.8 & 0.938 & 0.939 & 0.938 & 0.938 & 0.938 & 0.938 \\ 
5 & 88.2 & 0.882 & 0.885 & 0.882 & 0.882 & 0.882 & 0.882 \\ 
6 & 90.5 & 0.905 & 0.907 & 0.905 & 0.905 & 0.905 & 0.905 \\ 
7 & 83.5 & 0.835 & 0.834 & 0.835 & 0.835 & 0.835 & 0.834 \\ 
\bottomrule
\end{tabular}
\label{table:label generator performance on OL set, after pretraining - stronger teacher}
\end{table}

\begin{table}[H]
\caption{Macro F1-score the label generator (Fashion MNIST) obtains on each class}
\begin{tabular}{c c c c c c c c}
\toprule
\textbf{Classes} & \textbf{T-shirt} & \textbf{Trouser} & \textbf{Pullover} & \textbf{Dress} & \textbf{Coat} & \textbf{Sandal} & \textbf{Shirt} \\ 
\midrule
2 & 0.984 & 0.984 &  &  &  &  &  \\ 
3 & 0.950 & 0.984 & 0.957 &  &  &  &  \\ 
4 & 0.918 & 0.973 & 0.945 & 0.917 &  &  &  \\ 
5 & 0.914 & 0.973 & 0.822 & 0.889 & 0.814 &  & \\ 
6 & 0.902 & 0.966 & 0.843 & 0.877 & 0.847 & 0.996 &  \\ 
7 & 0.808 & 0.971 & 0.779 & 0.871 & 0.771 & 0.995 & 0.638 \\
\bottomrule
\end{tabular}
\label{table:label generator performance on each class, after pretraining - stronger teacher}
\end{table}

\begin{figure}[H]
\centering
\includegraphics[width=1.0\linewidth]{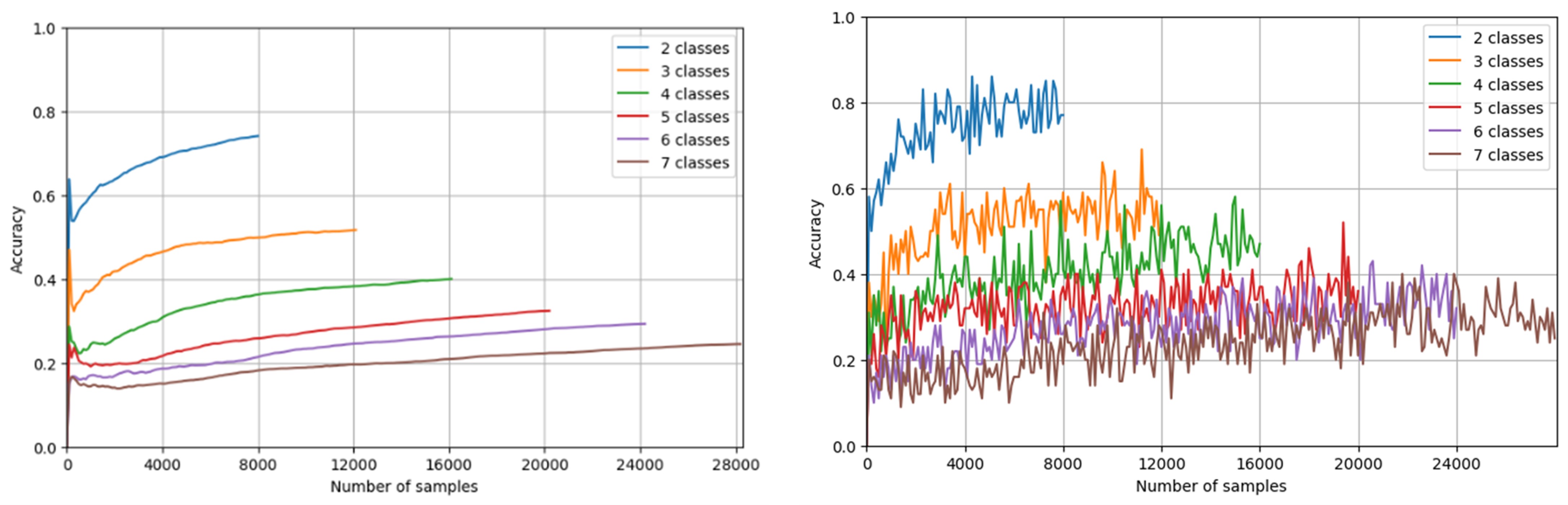}
\caption{Training process of the OL model (Facial Emotion)}
\label{fig:training process of OL model - stronger teacher}
\end{figure}

\begin{figure}[H]
\centering
\includegraphics[width=1.0\linewidth]{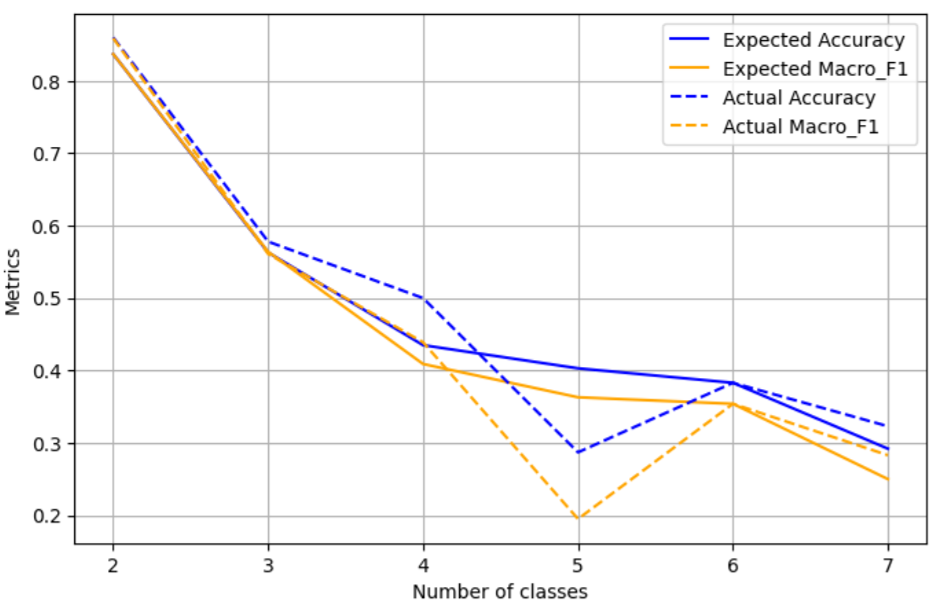}
\caption{Final performance of OL model (Facial Emotion)}
\label{fig:final performance of OL model - stronger teacher}
\end{figure}

In comparison with "weak teacher" (in \hyperref[fig:traing process of OL model - 1]{Fig.~\ref*{fig:traing process of OL model - 1}}). \hyperref[fig:training process of OL model - stronger teacher]{Fig.~\ref*{fig:training process of OL model - stronger teacher}} shows that the actual training could outperform the expected training if we look at the position where each line ends. As well, if we look closely at how the OL model performs on each class, we will see it has made better actual classifications. Finally, \hyperref[fig:final performance of OL model - stronger teacher]{Fig.~\ref*{fig:final performance of OL model - stronger teacher}} proves that by using a stable label generator model, the actual performance of the OL model is very close to its expected performance (i.e., less offset).

\begin{table}[H]
\caption{Macro F1-score the OL model (Facial Emotion) obtains on each class}
\begin{tabular}{@{}cccccccccccccccc@{}}
\toprule
\textbf{Classes} & \textbf{Angry} & \textbf{Disgust} & \textbf{Fear} & \textbf{Happy} & \textbf{Sad} & \textbf{Surprise} & \textbf{Neutral} \\  
\midrule
\multicolumn{8}{c}{\textbf{Expected}} \\
\midrule
2 & 0.836 & 0.838 &  &  &  &  &  \\ 
3 & 0.350 & 0.787 & 0.554 &  &  &  &  \\ 
4 & 0.122 & 0.668 & 0.456 & 0.390 &  &  &  \\ 
5 & 0.353 & 0.687 & 0.094 & 0.378 & 0.306 &  & \\ 
6 & 0.155 & 0.543 & 0.254 & 0.444 & 0.326 & 0.403 &  \\ 
7 & 0.169 & 0.618 & 0.177 & 0.313 & 0.304 & 0.020 & 0.149 \\ 
\midrule
\multicolumn{8}{c}{\textbf{Actual}} \\
\midrule
2 & 0.868 & 0.851 &  &  &  &  &  \\ 
3 & 0.393 & 0.760 & 0.529 &  &  &  &  \\ 
4 & 0.061 & 0.712 & 0.466 & 0.519 &  &  &  \\ 
5 & 0 & 0.470 & 0.30 & 0.003 & 0.201 &  & \\ 
6 & 0.104 & 0.573 & 0.193 & 0.423 & 0.359 & 0.470 &  \\ 
7 & 0.003 & 0.555 & 0.224 & 0.333 & 0.302 & 0.443 & 0.118 \\ 
\bottomrule
\end{tabular}
\label{table:OL model performance on each class - stronger teacher}
\end{table}

\section{Discussion}\label{discussion}
The comparison between using a weak label generator and a strong label generator is that the weak one does not maintain a stable performance across all classes. If we focus on the case of using the balanced OL dataset, \hyperref[table:label generator performance on each class, after pretraining]{Table.~\ref*{table:label generator performance on each class, after pretraining}} tells us that when the number of classes to use increases to 7, the model performed highly biased on different classes, which that the F1-score on classes “Disgust”, “Surprise” and “Happy” (0.822, 0.69 and 0.61) is much higher than the F1-score it obtained on other classes. This results in, when the label generator starts generating labels for the online-learning model, these three classes are more likely to be predicted by the label generator model, and the corresponding labels (based on the causal relationship) “Trouser”, “Sandal” and “Dress” appeared much more frequantly than other classes in Fashion MNIST. This is why the offset between the expected performance and the actual performance (in \hyperref[fig:final performance of OL model - 1]{Fig.~\ref*{fig:final performance of OL model - 1}}) becomes more serious when it crosses over the bound of binary classifications.

A more underlying observation is that, within such training process, there are four situations that can happen (\hyperref[table:four situations]{Table.~\ref*{table:four situations}}), where we believe cases 1 and 4 were the ‘culprit’. ‘T’ means True class and ‘F’ means False class. In case 1, when the model predicted something correctly, but it received the ‘ground truth’ saying that it is not, this will lead to a very bad consequence that the model updates the weight in the wrong direction. Where in case 4, the model predicted something incorrectly and received the 'ground truth' saying that "you have made a correct prediction", this theoretically does not ‘pollute’ the model, but it will make the model converge very slowly because the model will never get corrected. 
\begin{table}[H]
\caption{Possible situations within OL model during training}
\centering
\begin{tabular}{cccc}
\toprule
\textbf{Case} & \textbf{Predicted} & \textbf{Received} & \textbf{Result} \\ 
\midrule
1 & T & F & Bad \\ 
2 & T & T & Fine \\
3 & F & T & Fine \\
4 & F & F & Bad \\
\bottomrule
\end{tabular}
\label{table:four situations}
\end{table}

To avoid such situations, we need to make the label generator stably provide ‘T’, which guides to either case 2 or 3. The results proved the feasibility of the proposed method that, assuming the teacher model can maintain a good level of performance across all classes it knows (the best case is that the teacher model is a perfect model), then the student model can eventually reach a level of performance as expected at some point (the maximum performance that the model can reach from its inherent learning capability).

For the field of edge ML, the main model can be semi-trained on edge devices or on the server. This will depend on how many initial examples there are available and how much Flash memory the edge device has. For example, if there are 10 classes then at least 10 examples will be needed for semi-training, and suppose the samples are 28*28*1 images then we need 30.625 KB to store these initial samples. In such cases, the classification is probable to be directly semi-trained on edge devices. However, if there are a lot of initial samples available where the edge device cannot store all of them, then semi-training somewhere else would be more recommended. The label generator model can be pre-trained on edge devices or on the server (the latter should be more usual). This will depend on how complex the task the model is dealing with is, then how many initial examples there are available and how much Flash memory the edge device has. For example, if the task is very basic/simple, which means the model can reach a relatively good performance (e.g. 90+\% accuracy) with very few initial examples (e.g. 100 examples, suppose there are 10 classes, then 10 examples per each class, 306.25 KB needed). In such cases, the label generator model is possible to be directly pre-trained on edge devices. However, the example just given is kind of ideal because a model generally needs to be trained with a lot of data to achieve good performance, especially if the model is dealing with tasks related to detecting human reactions.

We may notice that the causal relationships can be very flexible due to specific scenarios, and it can represent even more complicated knowledge. For example, $P \Rightarrow Q \land W$ by pretraining one model we can obtain the labels for other two models; or $P\land Q \Rightarrow W$ by pretraining two models (assume the datasets for these two models are easy to collect) we can train the model whose label is very difficult or expensive to collect. On the other hand, KT does not have the restriction that the teacher model must be more complex than the student model, which typical knowledge distillation approaches do. Formally speaking, given two tasks, $P$ and $Q$, both settings $|\theta_T^{(P)}| \leq |\theta_S^{(Q)}|$ and $|\theta_T^{(P)}| \geq |\theta_S^{(Q)}|$ are allowed, where $|\theta_T^{(P)}|$ is the number of parameters the teacher model needs for stably handling the task $P$, and $|\theta_S^{(Q)}|$ is the number of parameters the student model needs to converge to a good performance for the task $Q$.

We end up with a question: “When does the OL model stop training and put into use?” In traditional ways, we evaluate how well a model is generalisable by using a test set to check its performance on unseen data. In our experiments, the label generator model can be evaluated in the traditional way because the online-learning dataset is essentially a test set to it. The interesting part is about the main model because every new sample it captures is unseen to this model; therefore, we can set a state variable to store some metrics of the model, such as accuracy then keep updating this variable (e.g. for every 1000 samples or 10k samples) until it exceeds a certain threshold (e.g. accuracy $>=$ 95\%).

\section{Conclusion}\label{conclusion}
In this paper, we aimed at exploring the problem “For truly Online Edge ML, how to get the label for future data”, which we believe is a fundamental problem in this field. We proposed a hybrid approach for co-training which allows the knowledge to be transformed between models, and we conducted simulation experiments to prove the feasibility of this method and also its effectiveness. The proposed method is suitable for the scenario where the labelled data for the teacher model is easy to gather, while the label for the student model is difficult and/or expensive to obtain. Knowledge Transformation is very flexible to various of environments since it lies on logic and can hence represent most knowledge in real life. We hope KT not only benefits the field of edge ML but can also inspire other fields in ML.

For future work, we aim to discover the way of generalising the proposed method, making it capable of adapting to arbitrary tasks rather than task-specific scenarios. The current study mainly focuses on image classification tasks; hence, exploring adapting the method to other ML tasks is needed. Furthermore, would it be possible to train both the teacher model and the student model simultaneously? This will be an intriguing topic to find out.  

\section*{Acknowledgements}
I would like to express my sincere gratitude to Prof. Anil Fernando for providing the supervision for this work. 

\nocite{*}
\bibliography{reference}

\end{document}